\pdfoutput=1

\documentclass[11pt]{article}

\usepackage[]{acl}

\usepackage{times}
\usepackage{latexsym}
\usepackage{paralist}
\usepackage{booktabs}
\usepackage{makecell}
\usepackage[T1]{fontenc}

\usepackage[utf8]{inputenc}
\usepackage{color}
\definecolor{DarkGreen}{RGB}{0,100,0}
\definecolor{DarkRed}{RGB}{139,0,0}
\definecolor{Orange}{RGB}{255,188,0}
\usepackage{graphicx}
\usepackage{ragged2e}
\usepackage{microtype}
\usepackage{amsmath}
\usepackage{amssymb}
\usepackage{arydshln}

\usepackage{latexsym}
\usepackage{times}
\usepackage{soul}
\usepackage{url}

\usepackage{graphicx}
\usepackage{amsthm}
\usepackage{amssymb}
\usepackage{amsmath}
\usepackage{multirow}
\usepackage{color}
\usepackage{booktabs}
\usepackage{algorithm}
\usepackage{algorithmic}
\usepackage{newfloat}
\usepackage{listings}
\usepackage{helvet}  
\usepackage{courier}  
\usepackage{graphicx} 
\usepackage{verbatim}
\usepackage{subfigure}
\usepackage{svg}
\usepackage{bm}
\usepackage{pifont}
\title{A Context-Aware Approach for Textual Adversarial Attack through {P}robability {D}ifference Guided {B}eam {S}earch}

\author{Huijun Liu, Jie Yu, Shasha Li, Jun Ma, Bin Ji \\
         College of Computer, National University of Defense Technology}

\begin{document}
\maketitle
\begin{abstract}
Textual adversarial attacks expose the vulnerabilities of text classifiers and can be used to improve their robustness.
Existing context-aware methods solely consider the gold label probability and use the greedy search when searching an attack path, often limiting the attack efficiency.
To tackle these issues, we propose \textbf{PDBS}, a context-aware textual adversarial attack model using \textbf{P}robability \textbf{D}ifference guided \textbf{B}eam \textbf{S}earch.
The probability difference is an overall consideration of all class label probabilities, and PDBS uses it to guide the selection of attack paths. In addition, PDBS uses the beam search to find a successful attack path, thus avoiding suffering from limited search space. 
Extensive experiments and human evaluation demonstrate that PDBS outperforms previous best models in a series of evaluation metrics, especially bringing up to a +19.5\% attack success rate. Ablation studies and qualitative analyses further confirm the efficiency of PDBS.

\end{abstract}

\section{Introduction}

In this paper, we explore Pre-trained Language Model (PLM) based textual adversarial attack, and focus on the score-based and non-targeted black-box setting. Previous work \cite{papernot2016, TextFooler} demonstrate that applying tiny perturbations to texts may fool a text classifier (a.k.a. target model) while these perturbations are usually human imperceptible, raising concerns on safety in reality. However, comprehensive research on textual adversarial attack helps to improve the robustness of target models \cite{Wallace2019}, thus attracting much research attention.

\begin{figure*}[t]
\centering
\includegraphics[width=0.85\textwidth]{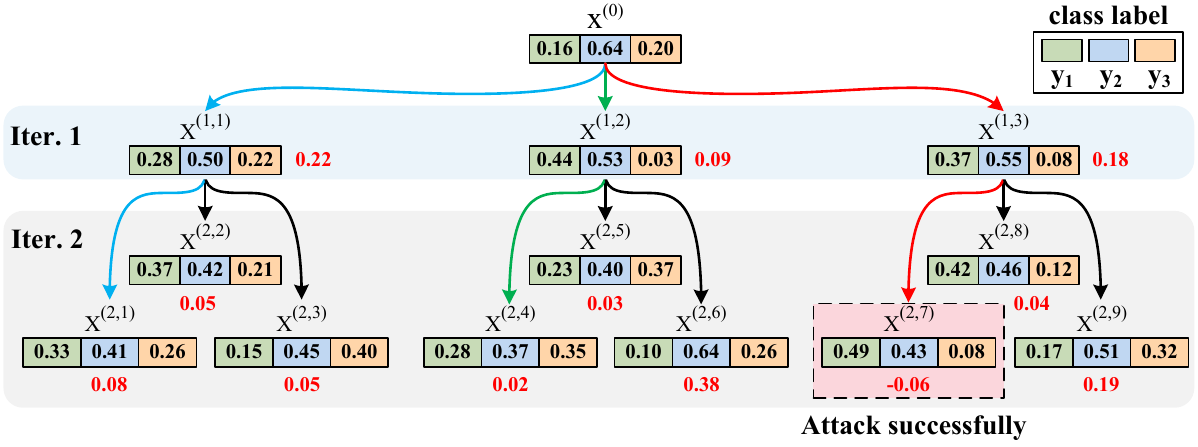} 
\caption{A textual attack example regarding the triple-class text classification, where $\{\textrm{y}_1, \textrm{y}_2$, $\textrm{y}_3\}$ are class labels. For clarity, we assume there are only three perturbations applied to one text sequence. $\textrm{x}^{(0)}$ is the input text sequence with gold label $\textrm{y}_2$ and Iter. 1 \& 2 are two attack iterations. The values in rectangles are probabilities of related class labels, which are predicted by the target model.  $\textrm{x}^{(i,j)}$ is a candidate example, where $i$ is the iteration number and $j$ denotes the $j$-$th$ candidate. The red value around $\textrm{x}^{(i,j)}$ is the \textbf{probability difference}, which is defined as the gold label ($\textrm{y}_2$) probability minus the maximum probability of the other labels (\{$\textrm{y}_1,\textrm{y}_3$\}). The blue path $\textrm{x}^{(0)} \rightarrow \textrm{x}^{(1,1)} \rightarrow \textrm{x}^{(2,1)}$ is the greedy search path when selecting the minimum gold label probability in each iteration. The green path $\textrm{x}^{(0)} \rightarrow \textrm{x}^{(1,2)} \rightarrow \textrm{x}^{(2,4)}$ is the greedy search path when selecting the minimum probability difference in each iteration. 
When using the probability difference guided beam search (set beam size to 2), we obtain a successful attack path -- the red path $\textrm{x}^{(0)} \rightarrow \textrm{x}^{(1,3)} \rightarrow \textrm{x}^{(2,7)}$, where $\textrm{x}^{(2,7)}$ fools the target model. However, using the gold label probability to guide the beam search (set size to 2) still leads to failure. 
}
\label{model2}
\end{figure*}

The most recent work introduces various PLMs, e.g., BERT \cite{BERT} and RoBERTa \cite{RoBERTa}, into the textual adversarial attack, since PLMs enable us to generate more contextual and fluent adversarial examples. Specifically, to generate adversarial examples, BERT-Attack \cite{BERTAttack} first finds the vulnerable tokens of the input text sequence through a Replace action and then uses BERT to generate substitutes for these tokens.  BAE \cite{BAE} uses Replace and Insert operations to iteratively mask vulnerable tokens of the input and then uses BERT to generate substitutes for the masked tokens. CLARE \cite{CLARE} uses Replace, Insert and Merge actions to mask tokens of the input text sequence and infills the masked tokens using substitutes generated by RoBERTa.
The score-based black-box setting assumes class label probabilities of a classifier are accessible. For example, the three label probabilities for input sequence $\textrm{x}^{(0)}$ are 0.16, 0.64, and 0.20 (see Figure \ref{model2}). 
Based on these scores, the above models formulate the task as a path search problem, and they propose to use the greedy search by selecting the minimum gold label probability in each attack iteration, such as the $\textrm{x}^{(0)} \rightarrow \textrm{x}^{(1,1)} \rightarrow \textrm{x}^{(2,1)}$.
However, we find that in multi-class ($\geq$ 3) classifications, solely considering the gold probability may be insufficient. For example, the $\textrm{x}^{(1,2)}$ is intuitively easier to attack successfully than the $\textrm{x}^{(1,1)}$, since the probability difference of the former (0.09) is less than the latter (0.22). 
But the greedy search selects $\textrm{x}^{(1,1)}$ when only considering the gold label probability. 

In addition, we demonstrate that the greedy search limits the search space, leading to attack failure in some cases. For example, in Figure \ref{model2} the greedy search fails anyway, whether using the gold class probability or the probability difference. However,  there is a successful attack path, namely $\textrm{x}^{(0)} \rightarrow \textrm{x}^{(1,3)} \rightarrow \textrm{x}^{(2,7)}$.

To tackle the above issues, we propose \textbf{PDBS}, a context-aware approach for textual adversarial attack through \textbf{P}robability \textbf{D}ifference guided \textbf{B}eam \textbf{S}earch.
Following CLARE \cite{CLARE}, PDBS uses three RoBERTa-based actions to obtain candidate adversarial examples, but it is different from CLARE in path search.
Specifically, we use the probability difference to guide the path search, where the probability difference is a trade-off between the probabilities of the gold label and the other labels.
In addition, we propose to use the beam search for the path search. Compared to the greedy search, the beam search provides multiple search channels, thus improving the attack success rate.
For example, in Figure \ref{model2} using the probability difference guided beam search enables us to obtain the successful path $\textrm{x}^{(0)} \rightarrow \textrm{x}^{(1,3)} \rightarrow \textrm{x}^{(2,7)}$.

We conduct experiments on Yelp, AG, MNLI, and QNLI to evaluate PDBS. Experimental results show that PDBS consistently outperforms previous state-of-the-art models in terms of A-rate, Mod, Sim, PPL, and overall GErr. Specifically, PDBS brings up to +19.5\% A-rate gains. In addition, the ablation studies ($\S$\ref{5.1}) validate the effectiveness of the probability difference and the beam search. 

In sum, we summarize the contributions as follows: (1) we propose the first textual attack model using probability difference to guide the attack path search; (2) we are the first to use the beam search in non-targeted textual adversarial attack; (3) our model achieves new state-of-the-art results across a set of benchmark datasets.

\section{Related Work}
Textual Adversarial attack has been widely studied. 
Previous work can be divided into two types: white-box \cite{papernot2016,Ebrahimi2018,Wallace2019,Meng2020} and black-box. The white-box work needs structures and parameters of target models, which are inaccessible in reality. By contrast, the black-box work only requires the easily accessible class probabilities, thus attracting much research attention. 

\subsection{Heuristic Rule-based Textual Adversarial Attack under Black-box Setting}
Early text adversarial attack models mainly use heuristic rule-based methods to generate candidate adversarial examples. Liang et al. \shortcite{Liang2018} use three actions to generate examples, but these examples are out of fluency and lose semantic information. To tackle this, recent work proposes more effective methods. Alzantot et al. \shortcite{Alzantot2018} use the genetic algorithm to find synonyms for word replacement. Ren et al. \shortcite{Ren2019} first obtain the word saliency and then greedily replace words with synonyms derived from WordNet \cite{WordNet}. Zang et al. \shortcite{Zang2020} use the particle swarm optimization algorithm to find synonyms having the same sememe of original words. Jin et al. \shortcite{TextFooler} first rank words and then greedily replace them with synonyms obtained by semantically enhanced embeddings \cite{mrkvsic2016}. Although these models achieve good results, they are context-unware, disabling them from generating contextual and fluent adversarial examples. By contrast, our model is based on the PLM, enabling us to avoid the above problems.

\subsection{PLM-based Textual Adversarial Attack under Black-box Setting}
To generate candidate adversarial examples in a context-aware manner, recent work proposes to use PLMs.
Li et al. \shortcite{BERTAttack} first use a Replace operation to find the vulnerable words based on the gold label probability, then it iteratively replaces the vulnerable words with synonyms generated by BERT \cite{BERT}.
Garg and Ramakrishnan \shortcite{BAE} propose a model similar to Li et al. \shortcite{BERTAttack}, but they add an Insert operation when finding vulnerable words. 
More recently, Li et al. \shortcite{CLARE} propose CLARE, which uses a mask-then-infill strategy to obtain candidate adversarial examples with RoBERTa \cite{RoBERTa}. 
The above three PLM-based models generate much more contextual and fluent adversarial examples than rule-based models.
However, we demonstrate there are two common flaws in them: (1) they only use the gold label probability to guide the attack path search, and (2) they use the greedy search, which limits the search space. Compared to them, our model considers all label probabilities (probability difference) and uses the beam search while also based on the PLM.

\section{Approach}
\subsection{Problem Definition}

In this paper, we focus on generating context-aware textual adversarial examples under the black-box and non-targeted settings. Textual adversarial example generation centers around a text classification model $f(\cdot)$, referred to as the target model. The black-box setting enables us to access probability outputs of $f(\cdot)$. 
We first define a text classification dataset $\mathcal{D}=\{(\textrm{x}^{(1)},\textrm{y}^{(1)}),(\textrm{x}^{(2)},\textrm{y}^{(2)}),…,(\textrm{x}^{(d)},\textrm{y}^{(d)})\}$ and the class label set $ \mathcal{Y} = \{y_1,y_2,…, y_\ell\}$, where $\textrm{x}^{(i)}$ is a text sequence, $\textrm{y}^{(i)}$ is the gold class label of $\textrm{x}^{(i)}$ and $\textrm{y}^{(i)} \in \mathcal{Y}$. Given a text-label pair $(\textrm{x},\textrm{y}) \in \mathcal{D}$ (assume $f(\textrm{x}) = \textrm{y}$), an adversarial example $\textrm{x}'$ is supposed to modify $\textrm{x}$ to fool the target model: $f(\textrm{x}') \neq \textrm{y}$ (non-targeted). In addition,  modifications made to $\textrm{x}'$ should be minimal, such that $\textrm{x}'$ should be close to $\textrm{x}$ and a human cannot percesive the modifications \cite{CLARE}. We achieve above constraints by restricting the similarity between $\textrm{x}'$ and $\textrm{x}$, namely $\textrm{sim}(\textrm{x}',\textrm{x}) \geq \epsilon$, where $\textrm{sim}(\cdot,\cdot)$ calculates the similarity and $\epsilon$ is a pre-set threshold. Following previous work \cite{CLARE,TextFooler}, we use neural networks as the $\textrm{sim}(\cdot,\cdot)$ to calculate the cosine similarity of text pairs in the embedding space.

\subsection{Mask-then-Infill Strategy}\label{3.2}
Li et al. \shortcite{CLARE} demonstrate that Pre-trained Language Model (PLM) helps to produce adversarial examples that are more fluent and closer to the original text, and they propose CLARE, which contains a Mask-then-Infill strategy. The strategy consists of three perturbation operations, i.e., Replace (R), Insert (I) and Merge (M). For a fair comparison, we also use the strategy in our model. But we propose a more effective way to select substitute tokens (see Eq.\ref{8}) and a more effective scoring function to calculate action scores (see Eq.\ref{9}).

Given the text-label pair $\textrm{(x,y)}$ and a position $i$, where $\textrm{x} = x_1x_2…x_n$ and $1\leq i \leq n$, the above three operations apply masks at the given position of $\textrm{x}$ using operation-specific methods.

The Replace obtains a masked sequence by replacing the token $x_i$ with the [mask] token.
\begin{equation}
\textrm{R}(i, \textrm{[mask]}) = x_1...x_{i-1}\textrm{[mask]}x_{i+1}...x_n
\end{equation}

The Insert obtains a masked sequence by inserting the [MASK] token between the token ${x}_i$ and ${x}_{i+1}$, increasing the sequence length by 1.
\begin{equation}
\textrm{I}(i, \textrm{[mask]}) = x_1...x_i\textrm{[mask]}x_{i+1}...x_n
\end{equation}

The Merge obtains a masked sequence by replacing the bigram ${x}_i{x}_{i+1}$ with the [MASK] token, decreasing the sequence length by 1.\footnote{For the last position (i.e., $n$), the operation just replaces the token ${x}_n$ with the [MASK] token.}
\begin{equation}
\textrm{M}(i, \textrm{[mask]}) = x_1...x_{i-1}\textrm{[mask]}x_{i+2}...x_n
\end{equation}

A total of $3n$ masked sequences can be obtained for \textrm{x}: 3 sequences for each position. For clarity, we use $\hat{\textrm{x}}$ to denote any one of the $3n$ sequences. 
\begin{equation}
{\hat{\textrm{x}} \in \{ \textrm{R}(i, \textrm{[mask]}), \textrm{I}(i, \textrm{[mask]}), \textrm{M}(i, \textrm{[mask]})\}_{i=1}^{n}}
\end{equation}

The \textbf{infilling action} first use a MLM to generate a substitute token set $\mathcal{Z}$ for the mask positon in $\hat{\textrm{x}}$, where $\mathcal{Z}$ is the vocabulary list $\mathcal{V}$ of the MLM initially. Then it selects a $z \in \mathcal{Z}$ to infill the [mask] position in $\hat{\textrm{x}}$. 
\begin{equation}
\textrm{infill}(\hat{\textrm{x}},\textrm{[mask]},z) = x_1…x_{j-1}zx_{j+1}…x_m
\end{equation} 
where we assume the [mask] token is at the position $j$ and the length of $\hat{\textrm{x}}$ is $m$.\footnote{The Insert and Merge change sequence lengths, so we use a variable $m$ to denote the length of the masked sequence $\hat{\textrm{x}}$.} 
We refer to the $x_1…x_{j-1}zx_{j+1}…x_m$ as a \textbf{candidate adversarial example}, which is denoted as  $\hat{\textrm{x}}_z$ for clarity.

To obtain high-quality adversarial examples, the infilling action only selects $z$ when satisfying the following constraints.
\begin{compactitem}
\item The probability of $z$ predicted by the MLM is larger than a threshold: $p_{\scriptsize{\textrm{MLM}}}(z|\hat{\textrm{x}}) > \alpha$
\item The similarity between $\hat{\textrm{x}}_z$ and $\textrm{x}$ is larger than a threshold: $\textrm{sim}(\hat{\textrm{x}}_z, $\textrm{x}$) > \beta$
\end{compactitem}

We use the $\alpha$ and $\beta$ values reported in {CLARE}: $\alpha = 5e^{-3}$, $\beta=0.7$.
We then obtain the substitute token set $\mathcal{Z}$ as follows.
\begin{equation}
{ \mathcal{Z}= \{z \in \mathcal{V}\}  \ s.t.   \
p_{\scriptsize{\textrm{MLM}}}(z|\hat{\textrm{x}}) > \alpha, \textrm{sim}(\hat{\textrm{x}}_{z}, \textrm{x}) > \beta}
\end{equation}

Previous work \cite{BERTAttack,BAE,CLARE} selects the $z \in \mathcal{Z}$ that most confuses the $f(\cdot)$ as the final substitute token for the infilling action, where
\begin{equation}
z = \underset{z \in \mathcal{Z}}{\arg  \min} \  p_{{f}(\cdot)}(\textrm{y}|\hat{\textrm{x}}_z)
\label{7}
\end{equation}

We observe that the above $z$ solely considers the probability of the gold class label $\textrm{y}$. 
However, we demonstrate that for multi-class classification tasks (\# class $\geq$ 3), taking the probabilities of all class labels into consideration is more effective for generating adversarial examples. Thus we select the $z \in \mathcal{Z}$ according to the probability difference, which is calculated by gold label probability minus the maximum probability of the other class labels. 
\begin{equation}
z = \underset{z \in \mathcal{Z}, \ \textrm{y}' \in \mathcal{Y}\setminus \{\textrm{y}\}}{\arg  \min} \  [p_{{f}(\cdot)}(\textrm{y}|\hat{\textrm{x}}_z) - p_{{f}(\cdot)}(\textrm{y}'|\hat{\textrm{x}}_z) ]
\label{8}
\end{equation}
where $\mathcal{Y}\setminus \{\textrm{y}\}$ denotes removing $\textrm{y}$ from $\mathcal{Y}$. For binary classifications, Eq.\ref{7} and Eq.\ref{8} make no difference.

In addition, the Eq.\ref{7} uses the greedy search since it solely selects the token that minimizes the gold label probability, which limits the search space. To tackle this, we propose to use the beam search. Specifically, we select the $z$ for $\mathcal{K}$ times and search an attack path through $\mathcal{K}$ channels (see $\S\ref{beam}$), where $\mathcal{K}$ is the beam size. 
For the $k\mbox{-}th$ of the $\mathcal{K}$ selections, we set the constraint of Eq.\ref{8} to  ($z \in \mathcal{Z}\setminus\{z^{(1)},z^{(2)},...,z^{(k-1)}\}, \textrm{y}' \in \mathcal{Y}\setminus \{\textrm{y}\}$), where $\{z^{(1)},z^{(2)},...,z^{(k-1)}\}$ is the set of selected tokens. Finally, we obtain $\mathcal{K}$ infilling actions for the masked sequence $\hat{\textrm{x}}$, where each action infills the [mask] position with a token from
$\{z^{(1)},z^{(2)},...,z^{(\mathcal{K})}\}$. Thus for the text sequence $\textrm{x}$ which has $3n$ masked sequences, we obtain a total of $(3*n*\mathcal{K})$ actions. For each action, we use the probability difference as its score.
\begin{equation}
s_{(\textrm{x, y})}(a) = \underset{\textrm{y}' \in  \mathcal{Y}\setminus \{\textrm{y}\}}\min [p_{f(\cdot)}(\textrm{y}|a(\textrm{x})) - p_{f(\cdot)}(\textrm{y}'|a(\textrm{x}))]
\label{9}
\end{equation}
where $a$ denotes the action. $a(\textrm{x})$ denotes applying $a$ to x, generating a candidate adversarial example. And the lower the score, the better the action.

\subsection{ Probability Difference Guided Beam Search}\label{beam}
We formulate the textual attacking as a path search problem and use probability difference to guide the beam search to find an attack path, where the destination example in the path can fool the target model, as the red path $\textrm{x}^{(0)} \rightarrow \textrm{x}^{(1,3)} \rightarrow \textrm{x}^{(2,7)}$ in Figure \ref{model2} shows.

Algorithm 1 shows the path search procedure using PDBS. We divide the total $T$ search iterations (Line 6-16) into two cases: the first iteration and the subsequent iterations. In the first iteration (i.e., $t=1$, Line 8-10), we start the beam search from the sequence $\textrm{x}^{(0)}$, and first obtain its top $\mathcal{K}$ best actions according to action scores (Line 9). Then we sort these actions (Line 17) and sequentially apply them to $\textrm{x}^{(0)}$ (Line 18-19). Each action generates a candidate adversarial example (Line 20), and if the example can fool $f(\cdot)$, it attacks successfully and we return it (Line 21-22). Else we regard the $\mathcal{K}$ examples generated by the $\mathcal{K}$ actions as perturbed text sequences and pass them to the next iteration. 

If we fail to attack in the first iteration, we continue the beam search in subsequent iterations (i.e., $t \geq 2$, Line 11-16). In the $t$-$th$ iteration, we conduct path search on the $\mathcal{K}$ perturbed text sequences generated in the $(t-1)$ iteration. For each of the $\mathcal{K}$ sequences, we first obtain its top $\mathcal{K}$ best actions (Line 13). Thus we can obtain a total of $\mathcal{K}^2$ actions. Next, we first sort the $\mathcal{K}^2$ actions (Line 17), and then we select the top $\mathcal{K}$ best actions and sequentially apply them to their texts (Line 18-20). 
Each action generates a candidate adversarial example (Line 20), and if the example can fool $f(\cdot)$, it attacks successfully, and we return it (Line 21-22). Else we regard the $\mathcal{K}$ examples generated by the $\mathcal{K}$ actions as perturbed text sequences and pass them to the $(t+1)$ iteration.

\begin{table}[H] 
\centering
\begin{tabular}{l}
\toprule
\textbf{Algorithm 1} Adversarial Attack by PDBS                            \\ \midrule
\resizebox{0.950\linewidth}{!}
{
\begin{tabular}[c]{@{}l@{}}
1: \textbf{Input:} text sequence \textrm{x}, gold class label \textrm{y},
\\ \quad \quad \quad \quad \ target model $f(\cdot)$ \\
2: \textbf{Output:} An adversarial example \\
3: \textbf{Initialization}: $\textrm{x}^{(0)} \leftarrow \textrm{x}$ \\
4: $\mathcal{K} \leftarrow \ $beam size \\
5:  $  T \leftarrow \ $iteration times \\
\\
6: \textbf{for} $1\leq t \leq T$ \textbf{do} \\
7:\ \ \  \quad $\mathcal{P}_2 \leftarrow \varnothing$  \\
8:\ \ \  \quad  \textbf{if} \ $t == 1$ \ \textbf{do} \\
\quad \quad  \  \ \  \textcolor[RGB]{73,153,54}{// start beam search only from $\textrm{x}^{(0)}$} \\ 
9:\ \ \  \quad \quad $\mathcal{P}_2 \leftarrow $ \ GetBestActions($\textrm{x}^{(0)}$) \\
10: \quad   \textbf{end if} \\
11: \quad   \textbf{if \ $t  \geq 2$ \ do} \\
\quad \quad \ \ \ \textcolor[RGB]{73,153,54}{// continue beam search with $\mathcal{K}$ sequences} \\ 
12: \quad  \quad \textbf{for} $1 \leq k \leq \mathcal{K}$ \textbf{do}\\
13:\ \ \quad \quad   \quad $\mathcal{P}^{(k)} \leftarrow$\ \ GetBestActions($\textrm{x}^{(t-1,k)}$) \\
14:\ \ \quad \quad   \quad $\mathcal{P}_2 \leftarrow \mathcal{P}_2 \cup \mathcal{P}^{(k)}$ \\
15:\ \ \quad   \quad \textbf{end for} \\
\quad \quad \quad \quad \textcolor[RGB]{73,153,54}{// $\mathcal{P}_2$ contains $\mathcal{K}^2$ action-text pairs} \\ 
16: \quad   \textbf{end if} \\
\\
17:\ \ \quad Sort $\mathcal{P}_2$ based on action scores ascendingly \\
18:\ \ \quad \textbf{for} $1 \leq k \leq \mathcal{K}$ \textbf{do} \\
\quad \quad \quad \textcolor[RGB]{73,153,54}{// solely select the top $\mathcal{K}$ best pairs from $\mathcal{P}_2$} \\ 
19:\ \ \quad \quad ($a$, $\bar{\textrm{x}}$) $\leftarrow$ the $k\mbox{-}th$ best action-text pair in $\mathcal{P}_2$ \\
20:\ \ \quad \quad $\textrm{x}^{(t,k)} \leftarrow $ Apply $a$ to $\bar{\textrm{x}}$ \\
21:\ \ \quad \quad \textbf{if} $f(\textrm{x}^{(t,k)}) \neq \textrm{y}$ \textbf{then} \textbf{return} $\textrm{x}^{(t,k)}$ \\
\quad \quad \quad \quad \textcolor[RGB]{73,153,54}{// attack successfully} \\ 
22:\ \ \quad \quad \textbf{end if}\\
23:\ \ \quad \textbf{end for} \\
24: \textbf{end for}\\
25: \textbf{return} None \\
\\
26: \textbf{def}\ GetBestActions($\bar{\textrm{x}}$)\ \textbf{do} \\
\quad \  \ \  \textcolor[RGB]{73,153,54}{// return the top $\mathcal{K}$ best action-text pairs for $\bar{\textrm{x}}$} \\ 
27:\ \ \quad $\mathcal{P}_1 \leftarrow \varnothing$ \\
28:\ \ \quad $\mathcal{A} \leftarrow$ all infilling actions of $\bar{\textrm{x}}$\\
\quad \quad \  \ \  \ \textcolor[RGB]{73,153,54}{// a total of (3*$|\bar{\textrm{x}}|$*$\mathcal{K}$) actions} \\ 
29:\ \ \quad {Sort $\mathcal{A}$ based on action scores ascendingly} \\
\quad \quad \ \  \ \ \textcolor[RGB]{73,153,54}{// the lower the score, the better the action} \\ 
30: \ \ \  \ \  \textbf{for} $1 \leq k \leq \mathcal{K}$ \textbf{do} \\
31: \quad  \ \  \quad $a \leftarrow $ the $k\mbox{-}th$ action in $\mathcal{A}$ \\
\quad \quad \quad \  \ \ \ \ \textcolor[RGB]{73,153,54}{// select the ${k}\mbox{-}th$ best action from $\mathcal{A}$} \\ 
32: \quad \quad \ \  $\mathcal{P}_1 \leftarrow \mathcal{P}_1\cup \{(a,\bar{\textrm{x}})\}$ \\
33: \quad \ \  \textbf{end for} \\
34: \quad \ \  \textbf{return} $\mathcal{P}_1$ \\
35: \textbf{end def} \\ 
\bottomrule
\end{tabular}}
\end{tabular}
\label{alg}
\end{table}

\section{Experiment}
\subsection{Datasets and Baselines}
\textbf{Datasets.} To evaluate the proposed PDBS, we conduct experiments on the following benchmark datasets regarding text classification and Natural Language Inference (NLI). 
 
\begin{compactitem}
\item \textbf{Yelp} \cite{Zhang2015}: a sentiment classification dataset from restaurant reviews, containing two classes: positive and negative.
\item \textbf{AG} \cite{Zhang2015}: a dataset classifying news articles to four classes: world, sports, business, and science/technology.
\item \textbf{MNLI} \cite{Williams2018broad}: a triple-class classification dataset regarding NLI. It is composed of premise-hypothesis pairs. Each pair is labeled with the relation between the premise and hypothesis. The relation set contains \textit{entailment}, \textit{contradiction} and \textit{neutral}.
\item \textbf{QNLI} \cite{Wang2018glue}: a binary classification dataset regarding NLI. Each instance is a question-answer pair, and is labeled whether the answer corresponds to the question.

\end{compactitem}

We summarize more dataset details in Table \ref{1111} \cite{CLARE}. Following previous work \cite{TextFooler,BERTAttack,CLARE}, we evaluate our model on a set of 1,000 examples.\footnote{For a fair comaprison, we actually use the exact 1,000 examples used in {CLARE} for each dataset. \url{https://github.com/cookielee77/CLARE}}
In addition, to comprehensively evaluate the PDBS, we conduct experiments on four additional datasets: DBpedia ontology dataset \cite{Zhang2015}, Stanford sentiment treebank \cite{Socher2013recursive}, Microsoft Research Paraphrase Corpus \cite{Dolan2005automatically}, and Quora Question Pairs \cite{Wang2018glue}. Limited by space, we report the results of these additional datasets in Appendix \ref{A}.

\begin{table}[htb] \small
\centering
\renewcommand\tabcolsep{4.5pt}
\begin{tabular}{lccrcc}
\toprule
 \textbf{Dataset}   & \multicolumn{1}{l}{\textbf{\# Class}} & \multicolumn{1}{l}{\textbf{Train}} & \multicolumn{1}{l}{\textbf{Test}} & \multicolumn{1}{l}{\textbf{Avg Len}} & \multicolumn{1}{l}{\textbf{Acc(\%)}} \\ \midrule
 {Yelp}      & 2                                       & 560K                               & 38.0K                               & 130                                  & 95.9                                 \\
{AG} & 4                                       & 120K                               & 7.6K                              &\ \ 46                                   & 95.0                                 \\ \midrule
 
 {MNLI}      & 3                                       & 392K                               & 9.8K                              & 23/11                                & 84.3                                 \\ 

 {QNLI}      & 2                                       & 105K                               & 5.4K                              & 11/31                                & 91.4                                 \\\bottomrule
\end{tabular}
\caption{\label{1111}Dataset details. The ``Acc(\%)'' denotes the target model’s accuracy on the original test set without adversarial attack.}
\end{table}

\begin{table*}[htb] \small
\centering
\renewcommand\tabcolsep{4.5pt}
\begin{tabular}{lccccccccccc}
\toprule
                        & \multicolumn{5}{c}{\textbf{Yelp}}                                                                                                                 &                               & \multicolumn{5}{c}{\textbf{AG}}                                                                                                                       \\ \midrule
\textbf{Model}          & \textbf{A-rate($\%$)\textcolor{red}{$\uparrow$}} & \textbf{Mod($\%$)\textcolor{red}{\textcolor{red}{\textcolor{red}{\textcolor{blue}{$\downarrow$}}}}} & \textbf{Sim\textcolor{red}{$\uparrow$}} & \textbf{PPL\textcolor{red}{\textcolor{red}{\textcolor{red}{\textcolor{blue}{$\downarrow$}}}}} & \textbf{GErr\textcolor{red}{\textcolor{red}{\textcolor{red}{\textcolor{blue}{$\downarrow$}}}}} & \multicolumn{1}{r}{\textbf{}} & \textbf{A-rate($\%$)\textcolor{red}{$\uparrow$}} & \textbf{Mod($\%$)\textcolor{red}{\textcolor{red}{\textcolor{red}{\textcolor{blue}{$\downarrow$}}}}} & \textbf{Sim\textcolor{red}{$\uparrow$}} & \textbf{PPL\textcolor{red}{\textcolor{red}{\textcolor{red}{\textcolor{blue}{$\downarrow$}}}}} & \textbf{GErr\textcolor{red}{\textcolor{red}{\textcolor{red}{\textcolor{blue}{$\downarrow$}}}}} \\ \midrule
TextFooler              & 77.0                             & 16.6                           & 0.70                   & 163.3                    & 1.23                      & \multicolumn{1}{r}{}          & 56.1                             & 23.3                           & 0.69                   & 331.3                    & 1.43                      \\
\multicolumn{1}{c}{+LM} & 34.0                             & 17.4                           & 0.73                   & 90.0                     & 1.21                      & \multicolumn{1}{r}{}          & 23.1                             & 21.9                           & 0.74                   & 144.6                    & 1.07                      \\
BERT-Attack             & 71.8                             & 10.7                           & 0.72                   & 90.8                     & 0.27                      & \multicolumn{1}{r}{}          & 63.4                             &\ \  7.9                            & 0.71                   & 90.6                     & 0.25                      \\
CLARE                   & 79.7                             & 10.3                           & 0.78                   & 83.5                     & 0.25                      & \multicolumn{1}{r}{}          & 79.1                             &\ \  6.1                            & 0.76                   & 86.0                     & 0.17                      \\ \midrule
PDBS                    & \textbf{99.2}                    & \textbf{\ \ 4.9}                   & \textbf{0.80}          & \textbf{64.2}            & \textbf{0.13}             & \multicolumn{1}{r}{\textbf{}} & \textbf{87.7}                    & \textbf{\ \ 6.0}                   & \textbf{0.78}          & \textbf{84.7}            & \textbf{0.07}             \\ \toprule 
\textbf{}               & \multicolumn{5}{c}{\textbf{MNLI}}                                                                                                                 & \textbf{}                     & \multicolumn{5}{c}{\textbf{QNLI}}                                                                                                                 \\ \midrule 
\textbf{Model}          & \textbf{A-rate($\%$)\textcolor{red}{$\uparrow$}} & \textbf{Mod($\%$)\textcolor{red}{\textcolor{red}{\textcolor{red}{\textcolor{blue}{$\downarrow$}}}}} & \textbf{Sim\textcolor{red}{$\uparrow$}} & \textbf{PPL\textcolor{red}{\textcolor{red}{\textcolor{red}{\textcolor{blue}{$\downarrow$}}}}} & \textbf{GErr\textcolor{red}{\textcolor{red}{\textcolor{red}{\textcolor{blue}{$\downarrow$}}}}} & \multicolumn{1}{r}{\textbf{}} & \textbf{A-rate($\%$)\textcolor{red}{$\uparrow$}} & \textbf{Mod($\%$)\textcolor{red}{\textcolor{red}{\textcolor{red}{\textcolor{blue}{$\downarrow$}}}}} & \textbf{Sim\textcolor{red}{$\uparrow$}} & \textbf{PPL\textcolor{red}{\textcolor{red}{\textcolor{red}{\textcolor{blue}{$\downarrow$}}}}} & \textbf{GErr\textcolor{red}{\textcolor{red}{\textcolor{red}{\textcolor{blue}{$\downarrow$}}}}} \\ \midrule
TextFooler              & 59.8                             & 13.8                           & 0.73                   & 161.5                    & 0.63                      &                               & 57.8                             & 16.9                           & 0.72                   & 164.4                    & 0.62                      \\
\multicolumn{1}{c}{+LM} & 32.3                             & 12.4                           & 0.77                   & 91.9                     & 0.50                      &                               & 29.2                             & 17.3                           & 0.75                   & 85.0                     & 0.42                      \\
BERT-Attack             & 82.7                             &\ \  8.4                            & 0.77                   & 86.7                     & 0.04                      &                               & 76.7                             & 13.3                           & 0.73                   & 86.5                     & 0.03                      \\
CLARE                   & 88.1                             &\ \  7.5                            & 0.82                   & 82.7                     & \textbf{0.02}             &                               & 83.8                             & 11.8                           & 0.78                   & 76.7                     & \textbf{0.01}             \\ \midrule
PDBS                    & \textbf{98.3}                    & \textbf{\ \ 4.5}                   & \textbf{0.83}          & \textbf{73.9}            & 0.07                      &                               & \textbf{92.6}                    & \textbf{\ \ 8.3}                   & \textbf{0.79}          & \textbf{67.1}            & 0.07                      \\ \bottomrule
\end{tabular}

\caption{\label{2}Performance comparisons across the four datasets. We use fine-tuned BERT models as target models.
Bold font values denote the best results for each metric. \textcolor{red}{$\uparrow$}(\textcolor{red}{\textcolor{red}{\textcolor{red}{\textcolor{blue}{$\downarrow$}}}}) indicates the higher (lower) the better.}
\end{table*}

\noindent \textbf{Baselines.} We compare our model with previous state-of-the-art models regarding black-box textual adversarial attack, including TextFooler \cite{TextFooler}, TextFooler+LM \cite{CLARE}, BERT-Attack \cite{BERTAttack} and CLARE \cite{CLARE}, where the first two are heuristic rule-based and the last two are PLM-based methods.\footnote{Following CLARE, we do not list BAE \cite{BAE} since it has a similar performance as BERTAttack.} 
We report more details of these baselines in Appendix \ref{B}.

\subsection{Implementation Details}
Following CLARE, we use a distilled version of RoBERT to infill the masked tokens. Across all the datasets, we set the iteration times $T$ to 10, the beam size $\mathcal{K}$ to 10. Moreover, we investigate model performance against various $T$ values in Appendix \ref{E}.
As for other hyper-parameters, we use just the same settings to CLARE.
 Following previous work \cite{BERTAttack,CLARE}, we use fine-tuned BERT models as the target models.\footnote{For a fair comparison, we actually use the fine-tuned BERT models used in CLARE.} 

In addition, we also conduct experiments on two additional target models, namely Word-LSTM \cite{Hochreiter1997long} and ESIM \cite{Chen2017enhanced}, where the former is specific for text classification, and the latter is specific for NLI. We report experimental results on the two additional models in Appendix \ref{C}.

\subsection{Automatic Evaluation Metrics}
Following CLARE, we use the Attack success rate (\textbf{A-rate}), Modification Rate (\textbf{Mod}), Textual similarity (\textbf{Sim}), Perplexity (\textbf{PPL}) and Grammar error (\textbf{GErr}) to comprehensively evaluate the model performance, where the last four are used to evaluate the quality of adversarial examples. We report more metric details in Appendix \ref{D}.

\subsection{Main Results}\label{4.4}

We summarize the performance results in Table \ref{2}. Obviously, PDBS successfully triggers classification errors in the target models across the four datasets.

Specifically, PDBS achieves the current state-of-the-art performance in terms of A-rate, Mod, Sim, and PPL across the four datasets. To be more precise, (1) PDBS delivers +8.6\% to +19.5\% A-rate gains compared to the previous best model, i.e., CLARE. These gains explicitly validate the excellent attack ability of PDBS. (2) PDBS consistently decreases the Mod scores, especially up to -5.4\% on Yelp, implying that the adversarial examples are more humanly imperceptible. (3) The consistent optimal Sim scores indicate that our adversarial examples are closer to the original texts than those generated by previous models. (4) PDBS decreases the PPL by -9.75 on average, especially -19.3 on Yelp. These lower PPL scores confirm that our adversarial examples are more fluent than others.

In addition, PDBS is also superior to the listed baselines in terms of overall GErr scores (0.085 \textit{v.s.} 0.113, averaged scores), demonstrating that our adversarial examples are grammatically correct.

We attribute these exciting performance gains to the fact that (1) the probability difference takes all class probabilities into consideration and is more effective than solely using the gold label probability; (2) the beam search provides more search channels, enabling our model to avoid suffering from limited search space caused by the greedy search.

We present case studies of adversarial examples in Appendix \ref{F}.

\subsection{Human Evaluation}
To further evaluate the quality of adversarial examples, we set up human evaluations on three criteria: {similarity} (\textbf{Sim.}), {fluency \& grammaticality} (\textbf{Flu. \& Gra.}), and {label consistency} (\textbf{Con.}). We choose three independent volunteers who are native English speakers. 
We randomly select 200 instances (original text, adversarial example, gold label) from AG and MNLI respectively, where the adversarial examples can fool our PDBS.  

We first ask the volunteers to score (from 0-1) the similarity of original-adversarial text pairs. And the higher the score, the better the similarity. Then we evaluate the fluency and grammaticality by asking volunteers to score (from 1-5) the mixed adversarial examples and original texts. And the higher the score, the better the fluency and grammaticality. As for label consistency, we ask volunteers to annotate adversarial examples and compare their annotations with the gold labels. We finally average the scores marked by the three volunteers.

Table \ref{3} shows the evaluation results. We can observe that: (1) the {Sim.} scores are 0.86 and 0.89, indicating that the adversarial examples maintain the most semantics of the original texts, which preliminarily proves the validity of our model. (2) the {Flu. \& Gra.} scores of the adversarial examples are close to the original ones (3.6 \textit{v.s.} 4.1 and 3.7 \textit{v.s.} 3.7), implying that the adversarial examples have good fluency and satisfy the grammatical rules on the whole. (3) the {Con.} scores are 0.77 and 0.84, revealing that the majority of the adversarial examples are humanly imperceptible. 

\begin{table}[thb] \small
\centering
\renewcommand\tabcolsep{5.5pt}
\begin{tabular}{llccc}
\toprule
\multicolumn{2}{c}{\textbf{Dataset}}            & \multicolumn{1}{l}{\textbf{Sim.\textcolor{red}{$\uparrow$}}} & \multicolumn{1}{l}{\textbf{Flu. \& Gra.\textcolor{red}{$\uparrow$}}} & \multicolumn{1}{l}{\textbf{Con.\textcolor{red}{$\uparrow$}}} \\ \midrule
\multirow{2}{*}{{AG}} & Original    & \multirow{2}{*}{0.86}                             & 4.1                                                              & \multirow{2}{*}{0.77}                                    \\
                                  & Adversarial &                                                   & 3.6                                                              &                                                          \\ \midrule
\multirow{2}{*}{{MNLI}}    & Original    & \multirow{2}{*}{0.89}                             & 4.0                                                              & \multirow{2}{*}{0.84}                                    \\
                                  & Adversarial &                                                   & 3.7                                                              &                                                          \\ \bottomrule
\end{tabular}
\caption{\label{3}Human evaluation results. }
\end{table}

In addition, the results on AG are worse than the ones on MNLI. We attribute it to the fact that the adversarial examples of AG have an overall higher modification rate than MNLI (6.2\% \textit{v.s.} 4.4\%).\footnote{Since we solely select a part of the two datasets, thus the two Mod scores are not the same to the ones in Table \ref{2}.}  

\section{Analysis}
We first conduct ablation studies on the probability difference and the beam search ($\S5.1$). Then we
examine the transferability of adversarial examples ($\S5.2$). At last, we explore using PDBS to improve the robustness of the target models ($\S5.3$).

\subsection{Ablation Study} \label{5.1}
\textbf{Ablation on the Probability Difference.}
We report the ablation results on the probability difference (\textbf{P\&D}) in Table \ref{4}. We use the \texttt{-P\&D} to denotes ablating the P\&D from PDBS, instead we use the gold label probability to select substitutes (see Eq. \ref{7}, also select $\mathcal{K}$ times) and calculate the action score.
\begin{equation}
s_{(\textrm{x, y})}(a) = p_{f(\cdot)}(\textrm{y}|a(\textrm{x})) 
\end{equation}

The ablation results show that the P\&D significantly promotes the A-rate, delivering +16.3\% and +15.6\% gains on MNLI and AG respectively, but the four quality evaluation metrics become worse in current scenario.\footnote{Since the probability difference and the gold label probability make no difference in binary classifications (see $\S$\ref{3.2}), we do not conduct ablations on Yelp and QNLI. }\footnote{In both datasets, the A-rate of \texttt{-P\&D} is inferior to CLARE, which we attribute to the fact that \texttt{-P\&D} and CLARE use different $T$ values. For a fair comparison, we conduct more experiments in Appendix \ref{G}.} 
We attribute the worse results to the fact that the \texttt{-P\&D} generates much fewer adversarial examples, thus making it unfair for comparison. For a fair comparison, we select a part of the 1000 test instances of the two datasets, where if the \texttt{-P\&D} can attack an instance successfully, then the instance is selected. Finally, 785 and 672 instances are selected for MNLI and AG, which are referred to as \textbf{Reduced MNLI} and \textbf{Reduced AG}, respectively. We report extended ablation results on the two reduced datasets in Table \ref{5}. We observe that (1) the two A-rates of PDBS are also 100\%; (2) the P\&D consistently benefits PDBS in terms of the four quality evaluation metrics.

We conclude that using the probability difference is more effective than using the gold label probability, especially in terms of the A-rate.

\begin{table}[thb] \small
\centering
\renewcommand\tabcolsep{2pt}
\begin{tabular}{lccccc}
\toprule
                    \multicolumn{6}{c}{\textbf{MNLI}}                                                                                                                                                                                                                     \\ \midrule
\textbf{Model}     & \multicolumn{1}{r}{\textbf{A-rate($\%$)\color{red}{$\uparrow$}}}  & \multicolumn{1}{r}{\textbf{Mod($\%$)\color{blue}{$\downarrow$}}} & \multicolumn{1}{r}{\textbf{Sim\color{red}{$\uparrow$}}} & \multicolumn{1}{r}{\textbf{PPL\color{blue}{$\downarrow$}}} & \multicolumn{1}{r}{\textbf{GErr\color{blue}{$\downarrow$}}} \\ \midrule
{PDBS}      & \textbf{98.3}                                        & 4.5                                                & 0.83                              & 73.9                                         & 0.07                                          \\ 
{\quad \texttt{-P\&D}} & 82.0                                                & \textbf{3.4}                                       & \textbf{0.84}                              & \textbf{71.7}                                & \textbf{0.06}                                 \\
\toprule 
                    \multicolumn{6}{c}{\textbf{AG}}                                                                                                                                                                                                                       \\ \midrule
\textbf{Model}     & \multicolumn{1}{r}{\textbf{A-rate($\%$)\color{red}{$\uparrow$}}} & \multicolumn{1}{r}{\textbf{Mod($\%$)\color{blue}{$\downarrow$}}} & \multicolumn{1}{r}{\textbf{Sim\color{red}{$\uparrow$}}} & \multicolumn{1}{r}{\textbf{PPL\color{blue}{$\downarrow$}}} & \multicolumn{1}{r}{\textbf{GErr\color{blue}{$\downarrow$}}} \\ \midrule
{PDBS}      & \textbf{87.7}                                        & 6.0                                                & 0.78                                       & 84.7                                         & 0.07                                          \\
{\quad \texttt{-P\&D}} & 72.1                                                 & \textbf{4.2}                                       & \textbf{0.79}                              & \textbf{79.2}                                & \textbf{0.05}                                 \\
 \bottomrule
\end{tabular}
\caption{\label{4}Ablation results on probability difference.}
\end{table}

\begin{table}[h] \small
\centering
\renewcommand\tabcolsep{2pt}
\begin{tabular}{lccccc}
\toprule
                    \multicolumn{6}{c}{\textbf{Reduced MNLI}}                                                                                                                 \\ \midrule
\textbf{Model}     & \textbf{A-rate($\%$)\color{red}{$\uparrow$}}  & \textbf{Mod($\%$)\color{blue}{$\downarrow$}} & \textbf{Sim\color{red}{$\uparrow$}} & \textbf{PPL\color{blue}{$\downarrow$}} & \textbf{GErr\color{blue}{$\downarrow$}} \\ \midrule
{PDBS}      & \textbf{100.0}                  & \textbf{3.2}                   & \textbf{0.85}          & \textbf{69.9}            & \textbf{0.04}             \\
{\quad \texttt{-P\&D}} & \textbf{100.0}                  & 3.4                            & 0.84                   & 71.7                     & 0.06                      \\
   \toprule
                    \multicolumn{6}{c}{\textbf{Reduced AG}}                                                                                                                   \\ \midrule
\textbf{Model}     & \textbf{A-rate($\%$)\color{red}{$\uparrow$}} & \textbf{Mod($\%$)\color{blue}{$\downarrow$}} & \textbf{Sim\color{red}{$\uparrow$}} & \textbf{PPL\color{blue}{$\downarrow$}} & \textbf{GErr\color{blue}{$\downarrow$}} \\ \midrule
{PDBS}      & \textbf{100.0}                  & \textbf{4.1}                   & \textbf{0.81}          & \textbf{73.6}            & \textbf{0.04}             \\
{\quad \texttt{-P\&D}} & \textbf{100.0}                  & 4.2                            & 0.79                   & 79.2                     & 0.05                      \\
 \bottomrule
\end{tabular}
\caption{\label{5}Extended ablation results on probability difference.}
\end{table}

\begin{figure}[t]
\centering
\includegraphics[width=0.38\textwidth]{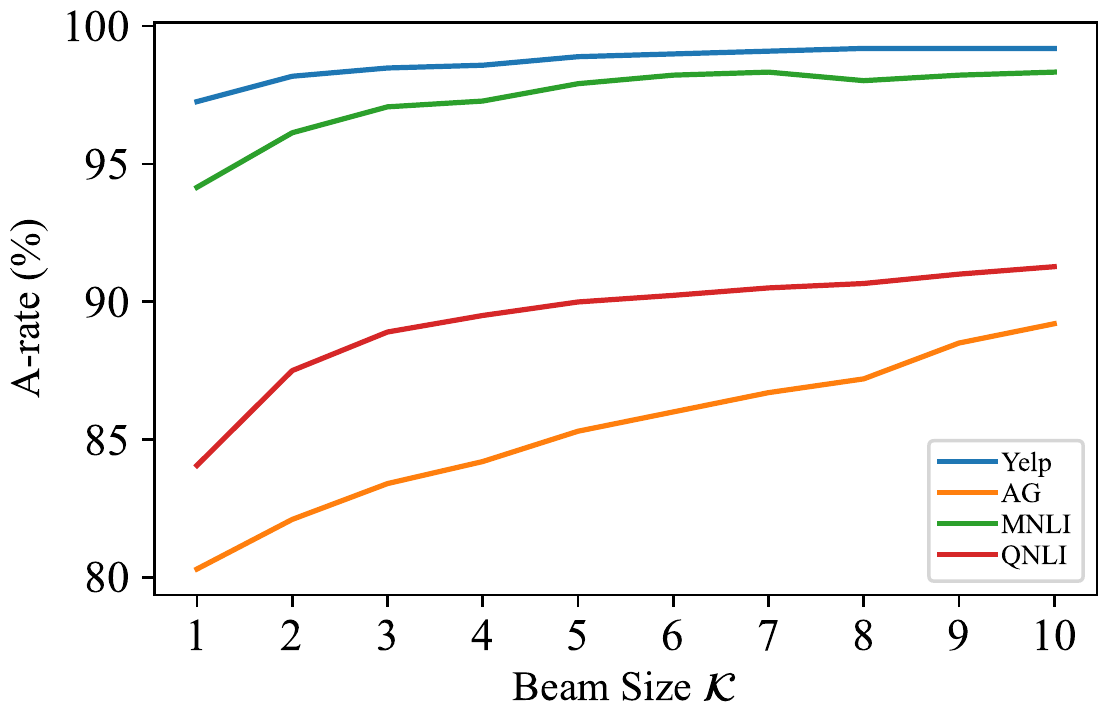} 
\caption{Ablation results on the beam search.
}
\label{model3}
\end{figure}

\noindent \textbf{Ablation on the Beam Search.}
We report the ablation results on the beam search in Figure \ref{model3}. To be specific, we set the beam size $\mathcal{K}$ from 1 to 10.
When $\mathcal{K}$=1, the beam search is the greedy search. We observe that (1) when the $\mathcal{K}$ increases, the A-rate generally increases across the four datasets; (2) compared to the greedy search ($\mathcal{K}$=1), the beam search ($\mathcal{K}$>1) consistently brings A-rate gains. For example, the beam search with $\mathcal{K}$=10 brings up to +17.62\% A-rate scores on AG. We attribute these to the fact that the beam search provides more search channels, thus improving the performance.

Additionally, we report the performance of the four quality evaluation metrics in Table \ref{6}. We observe that the $\mathcal{K}$ value poses tiny effects on the four metrics, revealing that a bigger $\mathcal{K}$ value can bring a better A-rate score while ensuring the high-quality of the adversarial examples.

Limited by hardware capacity, we cannot run experiments with a bigger $\mathcal{K}$ due to GPU out of memory with 24GB of GPU memory. We believe that it is possible to obtain a better A-rate with a bigger $\mathcal{K}$, especially on MNLI and AG.

\begin{table}[htb] \small
\centering
\renewcommand\tabcolsep{3pt}
\begin{tabular}{lcccc}
\toprule
                 & \textbf{Mod($\%$)} & \textbf{Sim} & \textbf{PPL} & \textbf{GErr} \\ \midrule
\textbf{Yelp}    & 5.22$\pm$0.17                 &0.80$\pm$0.001           &64.5$\pm$0.3           &0.133$\pm$0.002         \\
\textbf{AG} &5.94$\pm$0.08                 &0.79$\pm$0.004           &84.7$\pm$0.2           & 0.078$\pm$0.006            \\
\textbf{MNLI}    & 4.50$\pm$0.11          & 0.83$\pm$0.004  & 74.1$\pm$0.2    & 0.070$\pm$0.003   \\
\textbf{QNLI}    & 8.28$\pm$0.03                 & 0.79$\pm$0.001          & 66.8$\pm$0.3          &0.078$\pm$0.007           \\ \bottomrule
\end{tabular}
\caption{\label{6}Ablation results on the beam search when the $\mathcal{K}$ varing from 1 to 10. We report the averaged value of each metric, as well as its change intervals. 
}
\end{table}

\subsection{Transferability and Adversarial Training}
\textbf{Transferability.} The transferability is measured by if the adversarial examples generated with one target model can fool other target models directly \cite{Kurakin2016}.
For this purpose, we first collect the adversarial examples generated for Yelp and MNLI and then use other target models to predict their classes. Following previous work \cite{BERTAttack}, we select the Word-LSTM and ESIM as new target models for Yelp and MNLI, respectively. We report the results in Table \ref{7},  from with we can see that
there is a noticeable degree of transferability between models, and the transferability is higher in Yelp than in MNLI, which is consistent with the conclusion in \cite{TextFooler}.

\begin{table}[htb] \small
\centering
\begin{tabular}{llcc}
\toprule
\textbf{Dataset}               & \textbf{Model} & BERT & Word-LSTM \\ \midrule
\multirow{2}{*}{{Yelp}} & BERT           & \ \ 0.0  & 52.5      \\ 
                               & Word-LSTM      & 56.7 &\ \  0.0       \\ \toprule
\textbf{Dataset}               & \textbf{Model} & BERT & ESIM      \\ \midrule
\multirow{2}{*}{{MNLI}} & BERT           &\ \  0.0  & 65.1      \\ 
                               & ESIM           & 72.6 &\ \  0.0       \\ \bottomrule
\end{tabular}
\caption{\label{7}Transferability results. Row $i$ and column $j$ is the accuracy of adversarial examples generated for model $i$ evaluated on model $j$. The lower the score, the better the transferability.
}
\end{table}

\noindent \textbf{Adversarial Training.} We investigate if the adversarial examples can be used to improve the robustness of target models. For this purpose, we conduct a preliminary experiment on adversarial training by feeding the models both the original texts and the adversarial examples, where adversarial examples share the same labels as their original texts.

We collect the adversarial examples curated from the training sets of AG and MNLI and add them to the original training sets. We then use the mixed sets to fine-tune BERT from scratch and attack this adversarially fine-tuned model. The results in Table \ref{8} show that the A-rate and Mod after adversarial re-training become worse, indicating the greater difficulty of attacking. This reveals one of the potencies of our attack system: we can enhance the robustness of a model to future attacks by training it with the generated adversarial examples.

\begin{table}[htb]\small
\centering
\renewcommand\tabcolsep{2pt}
\begin{tabular}{llrrr}
\toprule
                                   &              & \textbf{Acc($\%$)\textcolor{red}{$\uparrow$}} & \textbf{A-rate($\%$)\textcolor{blue}{$\downarrow$}} & \textbf{Mod($\%$)\textcolor{red}{$\uparrow$}} \\ \midrule
\multirow{2}{*}{\textbf{AG}} & Original      & 95.0                         & 85.7                               & 6.0                          \\
                                  & +Adv.Training & 94.8(-0.2)                   & 66.1(-19.6)                        & 12.4(+6.4)                   \\ \midrule
\multirow{2}{*}{\textbf{MNLI}}    & Original      & 84.3                         & 98.3                               & 4.5                          \\
                                  & +Adv.Training & 84.0(-0.3)                   & 80.8(-17.5)                        & 10.2(+5.7)                   \\ \bottomrule
\end{tabular}
\caption{\label{8}Comparisons of original fine-tuning (``Original'') and adversarial fine-tuning (``+ Adv. Training'')  BERT models on AG and MNLI. }
\end{table}

\section{Conclusion}
In this work, we present PDBS, a PLM-based text adversarial attack model. PDBS uses probability difference to guide the beam search when seeking a successful attack path. Experimental results demonstrate that PDBS delivers +8.6\% to +19.5\% attack access rate, and is consistently superior to previous models in four quality evaluation metrics, indicating that the PDBS owns both efficient attack ability and the ability to generate high-quality adversarial examples. Qualitative analyses further show significant advantages of PDBS.

\bibliography{anthology,custom}
\clearpage

\appendix
\section{Appendix Results on Different Datasets}\label{A}
We include the results of DBpedia ontology dataset (DBpedia), Stanford sentiment treebank (SST-2), Microsoft Research Paraphrase Corpus (MRPC), and Quora Question Pairs (QQP) in this section. And the fine-tuned BERTs are the target models. Table \ref{9} summarizes some dataset details. And Table \ref{10} shows the results of these datasets. Compared with all baselines, PDBS achieves overall best performance, and it is consistent with our observation in $\S$\ref{4.4}.

\begin{table}[htb] \small 
\centering
\renewcommand\tabcolsep{4.5pt}
\begin{tabular}{lcrrcc}
\hline
\textbf{Dataset} & \multicolumn{1}{l}{\textbf{\#Classes}} & \multicolumn{1}{l}{\textbf{Train}} & \multicolumn{1}{l}{\textbf{Test}} & \multicolumn{1}{l}{\textbf{Avg Len}} & \multicolumn{1}{l}{\textbf{Acc(\%)}} \\ \hline
\textbf{DBpedia} & 14                                     & 560K                               & 70K                               & 55                                   & 99.3                                 \\
\textbf{SST-2}   & 2                                      & 67K                                & 0.9K                              & 10                                   & 92.3                                 \\ \hline
\textbf{MRPC}    & 2                                      & 3.6K                               & 1.7K                              & 23/23                                & 81.4                                 \\
\textbf{QQP}     & 2                                      & 363K                               & 40K                               & 13/13                                & 91.4                                 \\ \hline
\end{tabular}
\caption{\label{9}Dataset details. The “Acc(\%)”denotes the target model’s accuracy on the original test set without adversarial attack.}
\end{table}

\begin{table*}[htb] \small
\centering
\renewcommand\tabcolsep{4.5pt}
\begin{tabular}{lrrrrrlrrrrr}
\toprule 
                        & \multicolumn{5}{c}{\textbf{DBpedia}}                                                                                                             &                               & \multicolumn{5}{c}{\textbf{SST-2}}                                                                                                               \\ \hline
\textbf{Model}          & \textbf{A-rate($\%$)\textcolor{red}{$\uparrow$}} & \textbf{Mod($\%$)\textcolor{blue}{$\downarrow$}} & \textbf{Sim\textcolor{red}{$\uparrow$}} & \textbf{PPL\textcolor{blue}{$\downarrow$}} & \textbf{GErr\textcolor{blue}{$\downarrow$}} & \multicolumn{1}{r}{\textbf{}} & \textbf{A-rate($\%$)\textcolor{red}{$\uparrow$}} & \textbf{Mod($\%$)\textcolor{blue}{$\downarrow$}} & \textbf{Sim\textcolor{red}{$\uparrow$}} & \textbf{PPL\textcolor{blue}{$\downarrow$}} & \textbf{GErr\textcolor{blue}{$\downarrow$}} \\ \hline
TextFooler              & 56.2                            & 24.9                           & 0.68                   & 182.5                    & 1.88                      & \multicolumn{1}{r}{}          & 89.8                            & 14.9                           & 0.69                   & 227.7                    & 0.53                      \\
\multicolumn{1}{c}{+LM} & 20.1                            & 22.4                           & 0.70                   & 84.0                     & 1.22                      & \multicolumn{1}{r}{}          & 51.7                            & 18.3                           & 0.69                   & 137.5                    & 0.50                      \\
BERT-Attack             & 60.7                            & 9.1                            & 0.69                   & 57.8                     & 0.20                      & \multicolumn{1}{r}{}          & 87.8                            & 8.1                            & 0.67                   & 142.9                    & 0.03                      \\
CLARE                   & 65.8                            & 7.0                            & 0.73                   & 53.3                     & \textbf{-0.03}            & \multicolumn{1}{r}{}          & 97.8                            & 7.5                            & 0.75                   & 137.4                    & \textbf{0.01}             \\ \hline
PDBS                    & \textbf{89.4}                   & \textbf{5.8}                   & \textbf{0.80}          & \textbf{50.2}            & 0.01                      & \multicolumn{1}{r}{\textbf{}} & \textbf{98.6}                   & \textbf{5.4}                   & \textbf{0.78}          & \textbf{125.6}           & 0.04                      \\ \hline  \hline
\textbf{}               & \multicolumn{5}{c}{\textbf{MRPC}}                                                                                                                & \textbf{}                     & \multicolumn{5}{c}{\textbf{QQP}}                                                                                                                 \\ \hline
\textbf{Model}          & \textbf{A-rate($\%$)\textcolor{red}{$\uparrow$}} & \textbf{Mod($\%$)\textcolor{blue}{$\downarrow$}} & \textbf{Sim\textcolor{red}{$\uparrow$}} & \textbf{PPL\textcolor{blue}{$\downarrow$}} & \textbf{GErr\textcolor{blue}{$\downarrow$}} & \multicolumn{1}{r}{\textbf{}} & \textbf{A-rate($\%$)\textcolor{red}{$\uparrow$}} & \textbf{Mod($\%$)\textcolor{blue}{$\downarrow$}} & \textbf{Sim\textcolor{red}{$\uparrow$}} & \textbf{PPL\textcolor{blue}{$\downarrow$}} & \textbf{GErr\textcolor{blue}{$\downarrow$}} \\ \hline
TextFooler              & 24.5                            & 10.6                           & 0.75                   & 118.8                    & 0.35                      &                               & 16.2                            & 12.7                           & 0.74                   & 145.2                    & 0.61                      \\
\multicolumn{1}{c}{+LM} & 12.9                            & 9.5                            & 0.79                   & 71.0                     & 0.29                      &                               & 7.8                             & 12.9                           & 0.77                   & 78.8                     & 0.21                      \\
BERT-Attack             & 29.7                            & 13.5                           & 0.79                   & 74.6                     & 0.05                      &                               & 24.2                            & 11.3                           & 0.71                   & 78.0                     & 0.25                      \\
CLARE                   & 34.8                            & 9.1                            & 0.83                   & 69.5                     & 0.02                      &                               & 27.7                            & 10.2                           & 0.76                   & 74.8                     & 0.14                      \\ \hline
PDBS                    & \textbf{42.7}                   & \textbf{7.6}                   & \textbf{0.85}          & \textbf{66.7}            & \textbf{0.01}             &                               & \textbf{39.2}                   & \textbf{8.9}                   & \textbf{0.79}          & \textbf{70.2}            & \textbf{0.06}             \\ \bottomrule 
\end{tabular}
\caption{\label{10}Performance comparisons across other four datasets. We use fine-tuned BERT models as target models. Bold font values denote the best results for each metric. \textcolor{red}{$\uparrow$}(\textcolor{red}{\textcolor{red}{\textcolor{red}{\textcolor{blue}{$\downarrow$}}}}) indicates the higher (lower) the better.}
\end{table*}

\begin{table*}[htb] \small
\centering
\renewcommand\tabcolsep{4.5pt}
\begin{tabular}{lrrrrrrrrrrr}
\toprule
\multicolumn{12}{c}{\textbf{Word-LSTM}}                                                                                                                                                                                                                                                                                                              \\ \hline
               & \multicolumn{5}{c}{\textbf{Yelp}}                                                                                                                & \multicolumn{1}{l}{}          & \multicolumn{5}{c}{\textbf{AG}}                                                                                                                  \\ \hline
\textbf{Model}          & \textbf{A-rate($\%$)\textcolor{red}{$\uparrow$}} & \textbf{Mod($\%$)\textcolor{blue}{$\downarrow$}} & \textbf{Sim\textcolor{red}{$\uparrow$}} & \textbf{PPL\textcolor{blue}{$\downarrow$}} & \textbf{GErr\textcolor{blue}{$\downarrow$}} & \multicolumn{1}{r}{\textbf{}} & \textbf{A-rate($\%$)\textcolor{red}{$\uparrow$}} & \textbf{Mod($\%$)\textcolor{blue}{$\downarrow$}} & \textbf{Sim\textcolor{red}{$\uparrow$}} & \textbf{PPL\textcolor{blue}{$\downarrow$}} & \textbf{GErr\textcolor{blue}{$\downarrow$}} \\ \hline
TextFooler     & 97.8                            & 10.6                           & 0.79                   & 143.4                    & 1.17                      &                               & 95.8                            & 18.6                           & 0.63                   & 196.8                    & 1.35                      \\
BERT-Attack    & 98.8                            & 4.7                            & 0.81                   & 87.3                     & 0.25                      &                               & 97.2                            & 11.7                           & 0.67                   & 97.5                     & 0.38                      \\
CLARE          & 99.1                            & 4.5                            & 0.83                   & 82.7                     & 0.21           &                               & 97.7                            & 10.5                           & 0.72                   & 89.1                     & 0.23             \\ \hline
PDBS           & \textbf{99.7}                   & \textbf{3.6}                   & \textbf{0.84}          & \textbf{65.3}            & \textbf{0.15}                      & \textbf{}                     & \textbf{98.7}                   & \textbf{9.1}                   & \textbf{0.76}          & \textbf{80.2}            & \textbf{0.09}                      \\ \hline \hline
\multicolumn{12}{c}{\textbf{ESIM}}                                                                                                                                                                                                                                                                                                                   \\ \hline 
\textbf{}      & \multicolumn{5}{c}{\textbf{MNLI}}                                                                                                                & \multicolumn{1}{l}{\textbf{}} & \multicolumn{5}{c}{\textbf{QNLI}}                                                                                                                \\ \hline
\textbf{Model}          & \textbf{A-rate($\%$)\textcolor{red}{$\uparrow$}} & \textbf{Mod($\%$)\textcolor{blue}{$\downarrow$}} & \textbf{Sim\textcolor{red}{$\uparrow$}} & \textbf{PPL\textcolor{blue}{$\downarrow$}} & \textbf{GErr\textcolor{blue}{$\downarrow$}} & \multicolumn{1}{r}{\textbf{}} & \textbf{A-rate($\%$)\textcolor{red}{$\uparrow$}} & \textbf{Mod($\%$)\textcolor{blue}{$\downarrow$}} & \textbf{Sim\textcolor{red}{$\uparrow$}} & \textbf{PPL\textcolor{blue}{$\downarrow$}} & \textbf{GErr\textcolor{blue}{$\downarrow$}} \\ \hline
TextFooler     & 90.1                            & 14.5                           & 0.59                   & 182.4                    & 0.79                      &                               & 88.0                            & 18.4                           & 0.67                   & 182.6                    & 0.74                      \\
BERT-Attack    & 87.6                            & 19.9                           & 0.57                   & 107.8                    & 0.14                      &                               & 89.6                            & 15.7                           & 0.70                   & 89.9                     & 0.09                      \\
CLARE          & 90.5                            & 12.4                           & 0.63                   & 80.3                     & \textbf{0.05}                      &                               & 91.9                            & 10.3                           & 0.79                   & 80.1                     & \textbf{0.07}                      \\ \hline
PDBS           & \textbf{97.4}                   & \textbf{9.8}                   & \textbf{0.65}          & \textbf{74.3}            & \textbf{0.05}             &                               & \textbf{95.9}                   & \textbf{8.9}                   & \textbf{0.81}          & \textbf{67.9}            & 0.08             \\ \bottomrule
\end{tabular}

\caption{\label{11}Performance comparisons of attacking the two target models of Word-LSTM and ESIM and their corresponding two datasets. Word-LSTM corresponds to Yelp and AG datasets for text classification tasks, and ESIM corresponds to MNLI and QNLI datasets for natural language inference tasks. Bold font values denote the best results for each metric. \textcolor{red}{$\uparrow$}(\textcolor{red}{\textcolor{red}{\textcolor{red}{\textcolor{blue}{$\downarrow$}}}}) indicates the higher (lower) the better.}
\end{table*}

\begin{table}[thb] \small
\centering
\renewcommand\tabcolsep{2pt}
\begin{tabular}{lccccc}
\toprule
\multicolumn{6}{c}{\textbf{MNLI}}                                                                                                                                        \\ \hline
\textbf{Model}     & \textbf{A-rate($\%$)\color{red}{$\uparrow$}}  & \textbf{Mod($\%$)\color{blue}{$\downarrow$}} & \textbf{Sim\color{red}{$\uparrow$}} & \textbf{PPL\color{blue}{$\downarrow$}} & \textbf{GErr\color{blue}{$\downarrow$}} \\ \hline
CLARE                     & 88.1                          & 7.5                          & 0.82                   & 82.7                     & 0.02                      \\
PDBS                      & \textbf{98.9}                 & 4.8                          & 0.82                   & 77.6                     & 0.08                      \\
\multicolumn{1}{c}{\texttt{-P\&D}} & 93.2                          & \textbf{4.6}                 & \textbf{0.84}          & \textbf{74.9}            & \textbf{0.07}             \\ \hline \hline
\multicolumn{6}{c}{\textbf{AG}}                                                                                                                                          \\ \hline
\textbf{Model}     & \textbf{A-rate($\%$)\color{red}{$\uparrow$}}  & \textbf{Mod($\%$)\color{blue}{$\downarrow$}} & \textbf{Sim\color{red}{$\uparrow$}} & \textbf{PPL\color{blue}{$\downarrow$}} & \textbf{GErr\color{blue}{$\downarrow$}} \\ \hline
CLARE                     & 79.1                          & \textbf{6.1}                 & 0.76                   & 86.0                     & 0.17                      \\
PDBS                      & \textbf{93.5}                 & 9.2                          & 0.75                   & 87.1                     & 0.09                      \\
\multicolumn{1}{c}{\texttt{-P\&D}} & 88.1                          & 7.8                          & \textbf{0.76}          & \textbf{85.5}            & \textbf{0.08}             \\ \bottomrule
\end{tabular}
\caption{\label{T=20}Ablation results on probability difference when $T$=20.}
\end{table}

\section{Baselines Details}\label{B}
We compare our model with recent state-of-the-art word-level black-box adversarial models, which from heuristic rule-based and PLM-based two aspects, including:

\noindent\textbf{Heuristic Rule-based Adversarial Attack.}

\begin{compactitem}
\item \textbf{TextFooler}: a state-of-the-art heuristic rule-based model proposed by \cite{TextFooler}. It rank tokens and replaces tokens with their synonyms by using counter-fitting word embeddings \cite{mrkvsic2016} in a greedy way.
\item \textbf{TextFooler+LM}: an improved variant model of TextFooler, which is implemented by \cite{CLARE}. It builds on the token replacement of TextFooler, and adds a small size GPT-2 language model to filter out candidate tokens that do not fit in the context by calculated perplexity. 
\end{compactitem}

\noindent\textbf{PLM-based Adversarial Attack.}
We refer to models that generate candidates using a pertrained language model (PLM) as PLM-based models.
\begin{compactitem}
\item \textbf{BERT-Attack}: one of the earliest PLM-based models proposed by \cite{BERTAttack}. It ranks tokens and replaces tokens with their synonyms by using BERT\cite{BERT} in a greedy way. In addition, BAE\cite{BAE} is one of the earliest PLM-based models, but we do not include it since it has a similar performance to BERTAttack.
\item \textbf{CLARE}: a state-of-the-art PLM-based model by \cite{CLARE}. It masks tokens and greedily infill masked tokens generated from a distill version of RoBERTa \cite{RoBERTa}.
\end{compactitem}

\section{Details of the Performance of PDBS when Maximum Number of Iteration Times Changes}\label{E}
\begin{figure*}[h]
\centering
\includegraphics[width=0.5\textwidth]{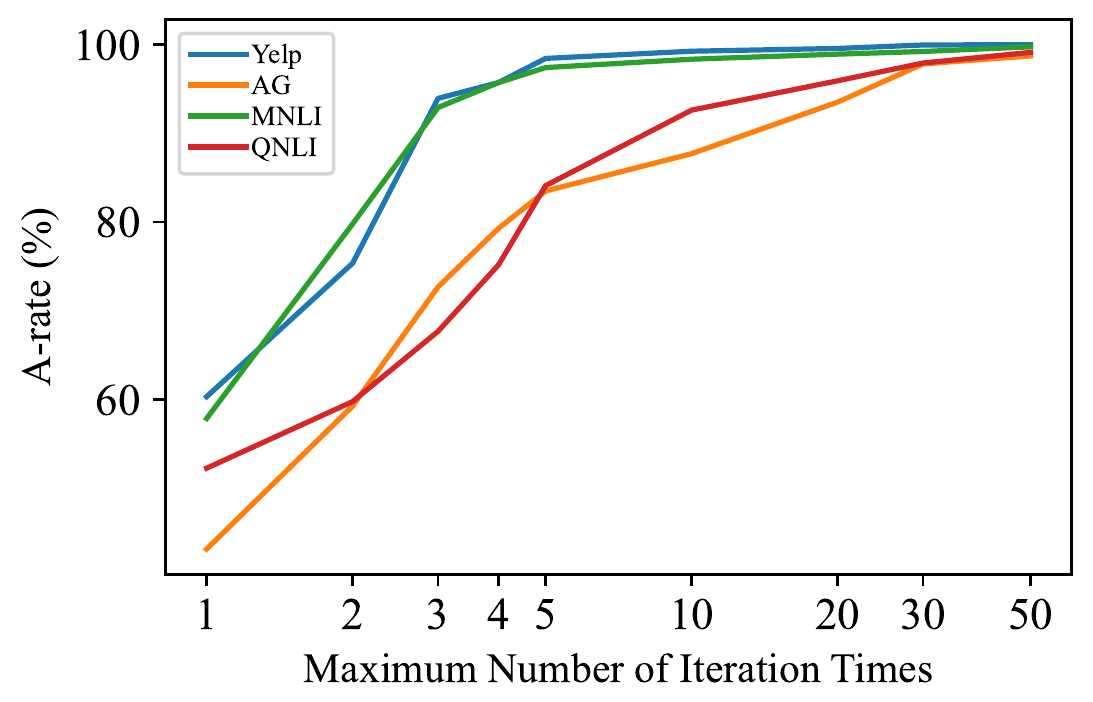} 
\caption{\label{Figure_A-rate}Performance of attack success rate (A-rate) on different datasets with different maximum number of iteration times $T$. The x-coordinate is on a log-2 scale. The larger the A-rate value, the better.
}
\label{model3}
\end{figure*}

\begin{figure*}[]
\centering
\includegraphics[width=1.0\textwidth]{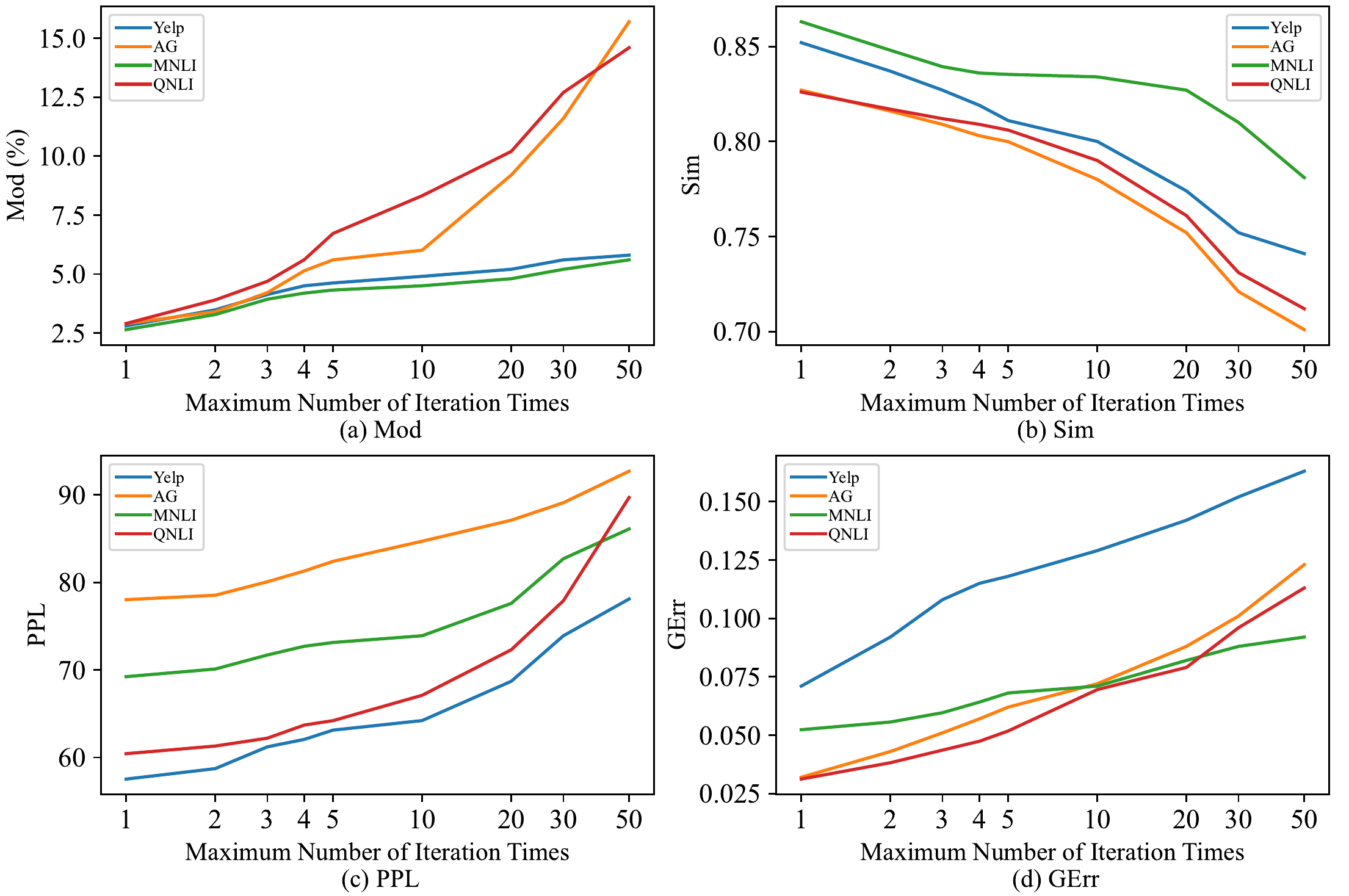} 
\caption{\label{Figure_Quality}Performance of adervsarial example quality, that is, modification rate (Mod),  textual similarity (Sim), perplexity (PPL), and grammar error (GErr) on different datasets with the different maximum number of iteration times $T$. The x-coordinate is on a log-2 scale. The larger the Sim, the better, and the smaller the Mod, PPL, and GErr, the better.}
\label{model3}
\end{figure*}

We analyze the performance of PDBS when the maximum number of iteration times $T$ changes. 
As shown in Figure \ref{Figure_A-rate} and \ref{Figure_Quality}, as the maximum number of iterations increases, the A-rate increases, but the quality of the generated adversarial examples decreases, as indicated by the rise in Mod, PPL, and GErr, as well as the drop in textual similarity. Note that the larger the values of A-rate and Sim, the better, and the smaller the values of Mod, PPL, and GErr, the better. To make a trade-off A-rate and the quality of adversarial examples, we select a uniform maximum number of iteration times $T$ = 10 for each dataset.
\section{Results on Different Target Models}\label{C}

To demonstrate that PDBS is also applicable to other target models, not limited to fine-tuned BERTs, we conduct experiments on two other target models, namely Word-LSTM and ESIM, where Word-LSTM is used for text classification tasks and ESIM is used for natural language inference tasks.
As seen in Table \ref{11}, PDBS is feasible for a wide range of target models. Similar to $\S$\ref{4.4}, PDBS outperforms previous best models, which can successfully trigger errors in Word-LSTM and ESIM with fewer modifications and higher quality adversarial examples.
Note that the results of TextFooler+LM are not listed because this section is not open in CLARE source code that proposes TextFooler+LM. The findings for TextFooler+LM are listed in the other sections since we directly take the results from the CLARE paper's relevant section.

\section{Automatic Evaluation Metrics Details}\label{D}
We evaluate the performance of attack models including their attack success rate and the quality of adversarial examples.

\noindent\textbf{Attack Success Rate (A-rate).}  
The percentage of adversarial examples which can trigger errors in the target model. It is the core metric to measure the success of the attacking method.

\noindent\textbf{Adversarial Example Quality.}
To evaluate the quality of the adversarial examples, we list four following evaluation metrics:

\begin{compactitem}
\item \textbf{Modification Rate (Mod)}: the percentage of the modified tokens to the text length. Following to \cite{CLARE}, each Replace and Insert operation represents one modified token; the Merge action is regarded to have modified one token if one of the two merged tokens is preserved, else it is considered to have modified two. 
\item \textbf{Textual Similarity (Sim)}: the semantic consistency between the adversarial example and the original text sequence. Following \cite{TextFooler, CLARE}, we use universal sentence encoder \cite{USE} to calculate the cosine semantic similarity. 
\item \textbf{Perplexity (PPL)}: the metric to measure the fluency of the adversarial examples \cite{Zang2020}. Following \cite{CLARE}, we calculate perplexity by using a small size of GPT-2 with a 50K-sized vocabulary \cite{GPT2}.
\item \textbf{Grammar Error (GErr)}: the metric calculated by the increase rate of grammatical error numbers of adversarial examples with the help of LanguageTool \cite{LanguageTool}.\footnote{\url{https://www.languagetool.org/}}
\end{compactitem}

\section{Ablation Study on Probability Difference when $T$=20}\label{G}
In MNLI and AG datasets, the attack success rate of \texttt{-P\&D} does not outperform CLARE, and we attribute the fact that \texttt{-P\&D} and CLARE employ different $T$ values which have a significant impact on the attack success rate (experimentally demonstrated in Appendix \ref{E}), where $T$=10 in \texttt{-P\&D} and $T$=20 in CLARE. However, some input sequences in \texttt{–P\&D} may require a $T$ value greater than 10 and less than 20 in order to be successfully attacked. In order to verify our idea, as well as for a fair comparison, we set $T$ in PDBS to 20 and conducted an ablation study on the probability difference. As Table \ref{T=20} shows, P\&D significantly promotes the A-rate, and using probability difference is more effective than using the gold label probability, which is consistent with our observation in \ref{5.1}. Additionally, it is demonstrated that the reason why \texttt{–P\&D} does not outperform CLARE in $\S$\ref{5.1} is that CLARE and \texttt{–P\&D} adopt different $T$ values.

\section{Case Study}\label{F}
We display some adversarial examples generated by CLARE and PDBS on Yelp, AG, MNLI and QNLI datasets in Table \ref{12}, \ref{13}, \ref{14}, and \ref{15} respectively.

\setlength{\belowdisplayskip}{1pt}

\begin{table*}[htb] \small
\centering
\renewcommand\tabcolsep{4.5pt}
\resizebox{1.0\linewidth}{!}
{
\begin{tabular}{l}
\toprule
\multicolumn{1}{c}{\textbf{Yelp Example 1}}                                                                                                                                                                                                                                                                                                                             \\ \hline
\textbf{Original Input (Prediction = Positive)}                                                                                                                                                                                                                                                                                                                          \\ \hline
\makecell[l]{This hotel is really nice, but it's mostly for the over 40 crowd. The restaurants are great, though. I stayed for 4 nights\\ and 
could not have been happier with the room accommodations or the friendliness of the staff. However, if you don't \\take Centrum and you want to go balls out Vegas crazy I recommend the Hard Rock.}                               \\ \hline
\textbf{CLARE (Prediction = Negative)}                                                                                                                                                                                                                                                                                                                                   \\ \hline
\makecell[l]{This hotel is really nice, but it’s mostly for the over \textcolor{red}{rated} 40 crowd. The restaurants are great, though. I stayed for 4 \\nights \textcolor{Orange}{flat}and could not have been happier with the room accommodations or \textcolor{Orange}{appreciate} the friendliness of the staff.\\ However, \textcolor{Orange}{but} if you don 't take Centrum and you want to go balls out Vegas crazy I recommend the Hard Rock.} \\ \hline
\textbf{PDBS (Prediction = Negative)}                                                                                                                                                                                                                                                                                                                                    \\ \hline
\makecell[l]{This hotel is really \textcolor{red}{being} nice, but it’s mostly for the over 40 crowd. The restaurants are great, though. I stayed for 4 \\nights and could not have been   happier with the room accommodations or the friendliness of the staff. However, if \\you don’t take Centrum and you want to go balls out Vegas crazy I recommend the Hard Rock.}                          \\ \hline \hline
\multicolumn{1}{c}{\textbf{Yelp Example 2}}                                                                                                                                                                                                                                                                                                                             \\ \hline
\textbf{Original Input (Prediction = Positive)}                                                                                                                                                                                                                                                                                                                          \\ \hline
\makecell[l]{Their food is always excellent, no matter what I’ve ordered (though I must say I love the Salmon salad with jalapeno \\dressing and the butternut squash ravioli with truffle oil). The firepit is relaxing. The music is fun. The service is \\sketchy ... touch and go. Don’t go if you’ve got a timeline.}                                                        \\ \hline
\textbf{CLARE (Prediction = Negative)}                                                                                                                                                                                                                                                                                                                                   \\ \hline
\makecell[l]{Their food is always excellent, no matter  what I’ve ordered (though I must say I love the Salmon salad with jalapeno   \\dressing and the \textcolor{red}{burnt} squash ravioli with   truffle oil). The firepit is relaxing. The music is fun. The service is   sketchy \\... \textcolor{red}{Stay} and \textcolor{Orange}{not} go. Don’t go if you’ve \textcolor{red}{established} a timeline.}\\ \hline
\textbf{PDBS (Prediction = Negative)}                                                                                                                                                                                                                                                                                                                                    \\ \hline
Their food \textcolor{Orange}{styling} is always excellent, no matter what I’ve ordered (though I must say I   love the Salmon salad with \\jalapeno dressing and the butternut squash ravioli   with truffle oil). The firepit is relaxing. The music is fun. The service \\is sketchy   ... touch and go. Don’t go if you’ve got a timeline.                                              \\ \bottomrule
\end{tabular}}
\caption{\label{12}Adversarial examples generated by CLARE and PDBS on Yelp dataset. \textbf{\textcolor{red}{Replace}},  \textbf{\textcolor{Orange}{Insert}} and \textbf{\textcolor{blue}{Merge}} are highlighted in \textcolor{red}{red}, \textcolor{Orange}{yellow} and \textcolor{blue}{blue}.
}
\end{table*}

\begin{table*}[htb] \small
\centering
\renewcommand\tabcolsep{4.5pt}
\resizebox{1\linewidth}{!}
{
\begin{tabular}{l}
\toprule
\multicolumn{1}{c}{\textbf{AG Example 1}}                                                                                                                                                               \\ \hline
\textbf{Original   Input (Prediction = Science\&Technology)}                                                                                                                                             \\ \hline
\makecell[l]{
ANOTHER VOICE Sugary drinks bad for you. Many studies have linked the consumption of   nondiet soda and fruit\\ juices with added sugars to obesity and attendant risks of diabetes.               } \\ \hline
\textbf{CLARE   (Prediction = World)}                                                                                                                                                                   \\ \hline
\makecell[l]{ANOTHER VOICE Sugary drinks bad for you. Many \textcolor{Orange}{European} studies have linked the consumption of nondiet soda \\and fruit juices with added sugars to obesity and attendant   risks of diabetes.}    \\ \hline
\textbf{PDBS   (Prediction = World)}                                                                                                                                                                    \\ \hline
\makecell[l]{
\textcolor{red}{RELATED} VOICE Sugary drinks bad for you. Many studies have linked the   consumption of nondiet soda and fruit \\juices with added sugars to obesity and attendant risks of diabetes. }                \\ \hline \hline
\multicolumn{1}{c}{\textbf{AG Example 2}}                                                                                                                                                               \\ \hline
\textbf{Original   Input (Prediction = Business)}                                                                                                                                                        \\ \hline
\makecell[l]{
Federated sales decline in August. A slow August snapped an eight-month winning streak for Federated Department   \\Stores Inc., which reported its first drop in sales since November.           }
\\ \hline
\textbf{CLARE   (Prediction = Science\&Technology)}                                                                                                                                                     \\ \hline
\makecell[l]{Federated sales decline in August. A slow August snapped an eight-month winning streak for Federated Department \\ \textcolor{Orange}{Apple} Stores Inc., which reported its first drop in sales since November.}       \\ \hline
\textbf{PDBS   (Prediction = Science\&Technology)}                                                                                                                                                      \\ \hline
\makecell[l]{Federated\textcolor{Orange}{'s} sales decline in August. A slow August snapped an eight-month winning streak for Federated Department \\Stores Inc., which reported its first drop in online sales since November.} \\ \bottomrule
\end{tabular}}
\caption{\label{13}Adversarial examples generated by CLARE and PDBS on AG dataset. \textbf{\textcolor{red}{Replace}},  \textbf{\textcolor{Orange}{Insert}} and \textbf{\textcolor{blue}{Merge}} are highlighted in \textcolor{red}{red}, \textcolor{Orange}{yellow} and \textcolor{blue}{blue}.
}

\end{table*}

\begin{table*}[htb] \small
\centering
\renewcommand\tabcolsep{4.5pt}
\resizebox{1\linewidth}{!}
{
\begin{tabular}{l}
\toprule
\multicolumn{1}{c}{\textbf{MNLI Example 1}}                                                                                                                                                                                                                                                                                                                                                                                                                                 \\ \hline
\textbf{Original   Input (Prediction = Contradiction)}                                                                                                                                                                                                                                                                                                                                                                                                                       \\ \hline
\begin{tabular}[c]{@{}l@{}}\makecell[l]{\textit{Premise}:   But I still don't see how he managed to prove his alibi, and yet go to the chemist's shop? Poirot stared at me in \\surprise.}\\    \textit{Hypothesis}: It's hard to see how he managed to prove his story is true.\end{tabular}                                                                                                                                                                                                          \\ \hline
\textbf{CLARE   (Prediction = Neutral)}                                                                                                                                                                                                                                                                                                                                                                                                                                     \\ \hline
\begin{tabular}[c]{@{}l@{}}\makecell[l]{\textit{Premise}:   But I \textcolor{red}{absolutely} don't  see how he managed to prove his alibi, and  \textcolor{red}{instead}    \textcolor{Orange}{purposefully} \textcolor{red}{wandered} to the chemist's  \\ shop? Poirot stared hard at me in surprise  \textcolor{Orange}{gaze}.}\\     \textit{Hypothesis}: It's hard to see how he managed to prove his story is true.\end{tabular}                                                                                                                                                                  \\ \hline
\textbf{PDBS   (Prediction = Neutral)}                                                                                                                                                                                                                                                                                                                                                                                                                                      \\ \hline
\begin{tabular}[c]{@{}l@{}}\makecell[l]{\textit{Premise}: But I still don't see how hard he managed to prove his alibi, and yet  \textcolor{red}{proceeded} to the chemist's shop? Poirot   stared \\ at me in surprise.}\\    \textit{Hypothesis}: It's hard to see how he managed to prove his story is true.\end{tabular}                                                                                                                                                                                          \\ \hline \hline
\multicolumn{1}{c}{\textbf{MNLI Example 2}}                                                                                                                                                                                                                                                                                                                                                                                                                                 \\ \hline
\textbf{Original   Input (Prediction = Contradiction)}                                                                                                                                                                                                                                                                                                                                                                                                                       \\ \hline
\begin{tabular}[c]{@{}l@{}}\makecell[l]{\textit{Premise}:  And then i had well you know i'd eat like at   lunch at work we'd have a TV or whatever and then um as my kids \\got older   and started you know recognizing what was going on and i thought this isn't   really very good.}\\     \textit{Hypothesis}: I'd watch TV at work during lunch, and as my children got older   and started understanding the shows, I realized \\that it is not good content   for them to be viewing.\end{tabular} \\ \hline
\textbf{CLARE   (Prediction = Neutral)}                                                                                                                                                                                                                                                                                                                                                                                                                                     \\ \hline
\begin{tabular}[c]{@{}l@{}}\makecell[l]{\textit{Premise}:   And then i had well you know i'd eat like at   lunch at work we'd have a TV or whatever and then  \textcolor{red}{...}   as my kids \\got older and started you know  \textcolor{red}{about} what   was going on and i thought  \textcolor{blue}{don}'t  \textcolor{red}{...} very good.}\\     \textit{Hypothesis}: I'd watch TV at work during lunch, and as my children got older and started understanding the shows, I realized\\ that it is not good content for them to be viewing.\end{tabular}            \\ \hline
\textbf{PDBS   (Prediction = Neutral)}                                                                                                                                                                                                                                                                                                                                                                                                                                      \\ \hline
\begin{tabular}[c]{@{}l@{}}\textit{Premise}:   And then i had well you know i'd eat like at   lunch at work we'd have a TV or whatever and then um as my kids \\got older and started you know  \textcolor{red}{...} what was going on and  i thought this isn't really very good.\\     \textit{Hypothesis}: I'd watch TV at work during lunch, and as my children got older and started understanding the shows, I realized\\ that it is not good content  for them to be viewing.\end{tabular}        \\ \bottomrule
\end{tabular}}
\caption{\label{14}Adversarial examples generated by CLARE and PDBS on MNLI dataset. \textbf{\textcolor{red}{Replace}},  \textbf{\textcolor{Orange}{Insert}} and \textbf{\textcolor{blue}{Merge}} are highlighted in \textcolor{red}{red}, \textcolor{Orange}{yellow} and \textcolor{blue}{blue}.
}
\end{table*}

\begin{table*}[htb] \small
\centering
\renewcommand\tabcolsep{4.5pt}
\resizebox{1\linewidth}{!}
{
\begin{tabular}{l}
\toprule
\multicolumn{1}{c}{\textbf{QNLI Example 1}}                                                                                                                                                                                                                                                                            \\ \hline
\textbf{Original   Input (Prediction = Not-Entailment)}                                                                                                                                                                                                                                                                 \\ \hline
\begin{tabular}[c]{@{}l@{}}\textit{Premise}:   Where can the public view the Royal Collection?\\    \makecell[l]{\textit{Hypothesis}: Unlike the palace and the castle, the purpose-built gallery is open continually and displays a changing selection\\ of items from the   collection.}\end{tabular}                                         \\ \hline
\textbf{CLARE   (Prediction = Entailment) }                                                                                                                                                                                                                                                            \\ \hline
\begin{tabular}[c]{@{}l@{}}\textit{Premise}:   Where can the public view the Royal Collection?\\    \makecell[l]{\textit{Hypothesis}: Unlike the palace and the castle, the \textcolor{red}{visitor}-built gallery is open continually \textcolor{Orange}{touring}  and \textcolor{red}{boasting} a changing \\selection of  items from the collection.}\end{tabular}                               \\ \hline
\textbf{PDBS   (Prediction = Entailment) }                                                                                                                                                                                                                                                             \\ \hline
\begin{tabular}[c]{@{}l@{}}\textit{Premise}:   Where can the public view the Royal Collection?\\     \makecell[l]{\textit{Hypothesis}:  Unlike the palace \textcolor{Orange}{grounds}  and the castle, the purpose-built \textcolor{Orange}{visitor}  gallery is open continually and displays a \\changing selection of items from the collection.}\end{tabular}                    \\ \hline \hline
\multicolumn{1}{c}{\textbf{QNLI Example 2}}                                                                                                                                                                                                                                                                            \\ \hline
\textbf{Original   Input (Prediction = Entailment)}                                                                                                                                                                                                                                                                     \\ \hline
\begin{tabular}[c]{@{}l@{}}\textit{Premise}:   Europe has a large market for what?\\     \makecell[l]{\textit{Hypothesis}: When Germany could no longer afford war payments, Wall Street   invested heavily in European debts to keep \\ the European economy afloat as a large consumer market for American mass-produced goods.}\end{tabular}  \\ \hline
\textbf{CLARE   (Prediction = Not-Entailment)}                                                                                                                                                                                                                                                                         \\ \hline
\begin{tabular}[c]{@{}l@{}}\textit{Premise}:   Europe has a large market for what?\\   \makecell[l]{ \textit{Hypothesis}: When Germany could no longer afford war payments, Wall Street invested heavily in European debts to keep \\ \textcolor{blue}{an} economy afloat as a large consumer base market for American mass-produced goods.}\end{tabular}       \\ \hline
\textbf{PDBS   (Prediction = Not-Entailment)}                                                                                                                                                                                                                                                                          \\ \hline
\begin{tabular}[c]{@{}l@{}}\textit{Premise}:   Europe has a large market for what?\\    \makecell[l]{\textit{Hypothesis}: When Germany could no longer afford war payments, Wall Street invested heavily in European debts to keep \\the European economy afloat as a large \textcolor{red}{dumping} market for American mass-produced goods.}\end{tabular} \\ \bottomrule
\end{tabular}}
\caption{\label{15}Adversarial examples generated by CLARE and PDBS on QNLI dataset. \textbf{\textcolor{red}{Replace}}, \textbf{\textcolor{Orange}{Insert}} and \textbf{\textcolor{blue}{Merge}} are highlighted in \textcolor{red}{red}, \textcolor{Orange}{yellow} and \textcolor{blue}{blue}.
}

\end{table*}

\end{document}